\documentclass[runningheads]{llncs}
\usepackage{times}
\usepackage{latexsym}
\usepackage{amssymb}
\usepackage{amsmath}
\usepackage{color}
\usepackage{tabularray}
\usepackage{enumerate}
\usepackage{graphicx}
\usepackage{paralist,bbding,pifont}
\usepackage[T1]{fontenc}
\newcommand{\figref}[1]{Figure ~\ref{#1}}
\usepackage{graphicx}
\usepackage{bbding}
\usepackage[misc]{ifsym}
\begin{document}
\title{A Novel ICD Coding Method Based on Associated and Hierarchical Code Description Distillation}
\titlerunning{A Novel ICD Coding Method AHDD}

\author{Bin Zhang\inst{1,2}\orcidID{0009-0007-0820-1965} \and
Junli Wang\inst{1,2}\textsuperscript{(\Letter)}\orcidID{0000-0002-7185-9731}}
\authorrunning{Bin Zhang, Junli Wang}

\institute{
Key Laboratory of Embedded System and Service Computing (Tongji University), \\ Ministry of Education, Shanghai 201804, China \and
National (Province-Ministry Joint) Collaborative Innovation Center \\ for Financial Network Security, Tongji University, Shanghai 201804, China \\
\email{2233009@tongji.edu.cn} \\
\email{junliwang@tongji.edu.cn}
}

\maketitle    
\begin{abstract}
ICD(International Classification of Diseases) coding involves assigning ICD codes to patients visit based on their medical notes. ICD coding is a challenging multilabel text classification problem due to noisy medical document inputs. Recent advancements in automated ICD coding have enhanced performance by integrating additional data and knowledge bases with the encoding of medical notes and codes. However, most of them ignore the code hierarchy, leading to improper code assignments. To address these problems, we propose a novel method based on associated and hierarchical code description distillation (AHDD) for better code representation learning and avoidance of improper code assignment.we utilize the code description and the hierarchical structure inherent to the ICD codes.
Therefore, in this paper, we leverage the code description and the hierarchical structure inherent to the ICD codes. 
The code description is also applied to aware the attention layer and output layer. Experimental results on the benchmark dataset show the superiority of the proposed method over several state-of-the-art baselines.

\end{abstract}

\section{Introduction}
The International Classification of Diseases (ICD), overseen by the World Health Organization, is a key coding system in healthcare, assigning codes to diagnostic and procedural data from patient visits. These codes standardize and systemize health information globally, aiding in epidemiological studies, service billing, and health trend monitoring \cite{epidemiological,monitor}. However, the manual nature of ICD coding presents significant challenges, including the training of medical experts and the difficulty of accurately assigning codes from a large set, as seen in the ICD-9 and ICD-10's extensive code lists. This manual process is costly, time-consuming, and prone to errors \cite{manual}, sparking growing interest in automated ICD coding in both industry and academia.

Viewing medical code prediction as a multi-label text classification challenge, a variety of machine learning-based methods have been developed and proposed. These include rule-based \cite{rule}, Support Vector Machine (SVM)-based \cite{svm}, and decision tree-based methods \cite{decision}. With the advent of deep learning in NLP tasks, a significant focus has been placed on learning deep representations of medical notes. This has been achieved using Recurrent Neural Networks (RNNs) \cite{LAAT,MSMN,2024AccurateAW}, Convolutional Neural Networks (CNNs) \cite{caml,multirescnn}, or Transformer \cite{Fusion} encoders, followed by multi-label classification for code prediction. 

Although effective, those methods do not target at the noisy text in medical note. Noisy text problem indicates that medical notes are lengthy and noisy, where only some key phrases are highly related to the coding task. Medical notes, typically authored by doctors and nurses, exhibit a range of writing styles. It is pointed out in \cite{noisy_distill} that approximately 10\% of words in a medical note are pertinent to the task of code assignment. 
Training models with such noisy text can lead to confusion about where to attention and result in erroneous decisions due to semantic deviations.

\begin{figure}[h]
	\centering
	\includegraphics[scale=0.60]{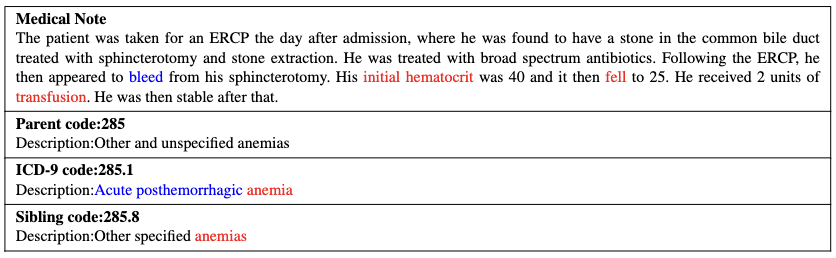} 
	\caption{An example of a medical note annotated with ICD-9 code "285.1", its parent code "285", and the sibling code "285.8". The words highlighted in red represent terms found in the descriptions of the child codes, aiding in the identification of key words within the medical note. Conversely, the words in blue denote terms that differ from the sibling code's description, serving to pinpoint more closely related words, thereby enhancing accuracy.}
	\label{Figure 1}
\end{figure}

To address the noisy text problem, a self-distillation learning mechanism is proposed to ensure the extracted shared representations focus on the long medical note's important part \cite{noisy_distill,2024CoRelationBA}. However, these methods overlook the code hierarchy, leading to improper code assignments, as the child codes of a parent are very similar clinically. To tackle this problem, we focus on the code descriptions and the hierarchical structure inherent to the ICD codes. Each ICD code typically comes with a concise description, and they are organized in a hierarchical manner. For example, as illustrated in \figref{Figure 1}, both ICD code 285.1 (Acute posthemorrhagic anemia) and ICD code 285.8 (Other specified anemias) are the children of ICD code 285 (Other and unspecified anemias). We can effectively identify key terms related to "\textit{anemia}" in code description, including "\textit{initial hematocrit}", "\textit{fell}" and "\textit{tranfusion}". These descriptions helps to streamline and clarify lengthy, noisy medical notes. However, since ICD code 285.1 and 285.8 are more clinically relevant, the model possibily assign both codes to a medical code. According to \cite{hierachy,2023KnowledgeMKFNMK}, it is unlikely to simultaneously assign all children codes of a parent to a medical code. To prevent inappropriate code assignments, we also capture important word "\textit{bleed}" in the document by the word "\textit{acute posthemorrhagic}" in code description different from sibling code.

Therefore, in this paper, we propose a novel method based on associated and hierarchical code description distillation to predict ICD codes. Specifically, the medical note, associated code and hierarchical code are fed into the shared encoder. We apply code description aware attention mechanism to get the label-specific representation. Then both associated code and hierarchical code applied to distill the medical note. Finally the code description aware output is applied for classification. The experimental results demonstrate that our proposed method surpasses several leading state-of-the-art models on a benchmark dataset.

The principal contributions of this paper are as follows:
\begin{compactitem}
\item  A novel method based on associated and hierarchical code description distillation (AHDD) is proposed to predict ICD codes. The code description is also applied to aware the attention layer and output layer. To the best of our knowledge, our work is the first attempt to use associated and hierarchical code descriptions for distilling medical notes.
\item The method is encoder-agnostic. And it neither needs extra text processing nor brings in too many parameters. Experimental results with a benchmark dataset demonstrate the effectiveness of the proposed method.
\item The extensive experiments about noisy text demonstrate that our proposed method can identifying key words within noisy and lengthy medical notes.
\end{compactitem}

\section{Related Work}
ICD coding has been a significant task garnering attention for decades. \cite{rule} utilized a rule-based methodology to extract essential snippets from medical notes, encoding them with ICD codes. \cite{svm} introduced an SVM classifier employing bag-of-words features. \cite{decision} attempted to identify key features using a decision tree-based approach. However, these methods did not attain satisfactory performance, primarily due to the challenge of extracting useful features from the complex and noisy medical notes.

With the popularity of neural networks, many researchers focused on using RNN CNN and Transformer models for ICD coding. \cite{multirescnn} used convolutional layers with different kernel sizes to extract relevant information for each code from source medical notes. \cite{LAAT} concentrated on identifying label-specific words in notes using an LSTM equipped with a custom label attention mechanism. \cite{transicd} was the first to propose a Transformer-based approach, achieving results on par with CNN-based models.

To further improve the performance, incorporating code descriptions and external knowledge had been explored. \cite{noisy_distill} trained a teacher network informed by label descriptions and modeled code co-occurrence using interactive shared attention. \cite{MSMN,2023KnowledgeMKFNMK,2024KnowledgebasedDP} harnessed code synonyms from a medical knowledge graph, adapting a multi-head attention mechanism. \cite{knowledge} integrated external Wikipedia knowledge and implemented a medical concept-driven attention. But all of them ignore the code hierarchy, which easily leads to  improper code assignments.

In terms of code hierarchy, \cite{hypercore} proposed a hyperbolic representation method to leverage the code hierarchy. \cite{LAAT,2023Rare-ICD,2024CoRelationBA}, introduced a hierarchical joint learning architecture, considering the hierarchical relationships among codes. However, these methods just use code hierarchy in output layer to avoid inappropriate code assignments. It is insufficient to capture the key words in medical note. Contrarily, we utilize the associated and hierarchical code description to capture the key word by distillation.

\begin{figure*}[h]
	\centering
	\includegraphics[scale=0.30]{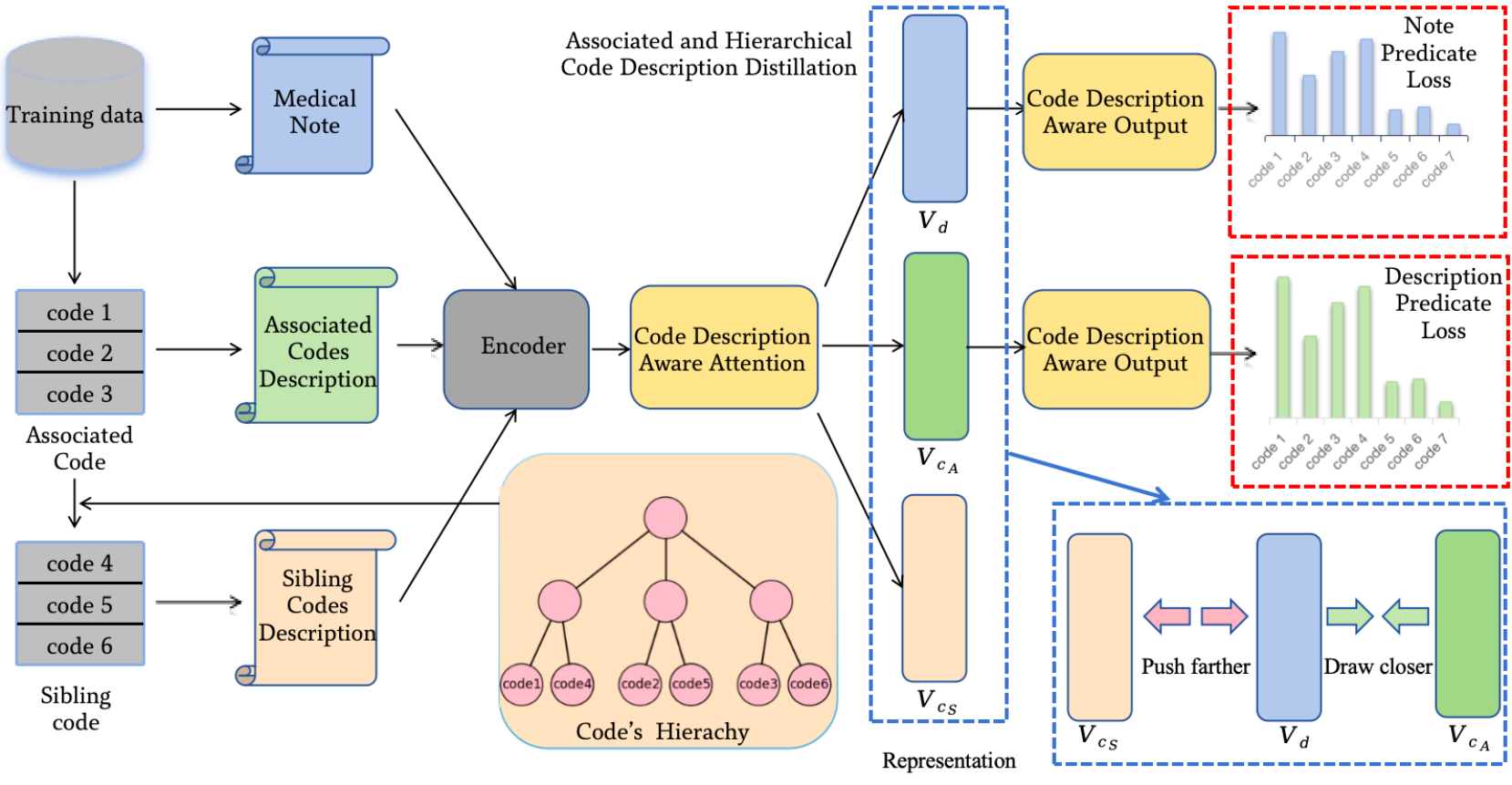} 
	\caption{The architecture of proposed \textbf{AHDD} method. \( V_d \), \( V_{c_A} \), and \( V_{c_S} \) represent the label-specific representation for the medical note, associated code, and sibling code, respectively. It is important to highlight that the Backbone Encoder can be implemented using various neural encoders. }
	\label{Figure 2} 
\end{figure*}

\section{Methods}
\textbf{ICD Coding:}
ICD coding represents a complex task of multi-label classification. Specifically, consider a medical note composed of \begin{math} N_d \end{math} tokens:
\begin{equation}
    d = \{t_1, t_2, \cdots, t_{N_d}\},
\end{equation}
and the medical code set \begin{math} L = \{l_1, l_2, \cdots, l_{N_L}\} \end{math}, where the size is denoted by \begin{math} N_L \end{math}. Additionally, each medical code is accompanied by a brief description, represented as \begin{math} c_i = \{t^c_1, t^c_2, \cdots, t^c_{N_{c_i}}\} \end{math}, where \begin{math} N_{c_i} \end{math} signifies the length of the code's description and \begin{math} i \in [1, N_L] \end{math}.

The multi-label classification is transformed into binary classification, where the objective is to assign a binary label \begin{math} l_i \in \{0,1\} \end{math} to each ICD code within the label space \begin{math} L \end{math}. Here, the label 1 indicates a positive association with an ICD-identified disease or procedure.

The overall architecture of the proposed method (AHDD) is shown in \figref{Figure 2}, which consists of four parts: (1) Encoder Layer which generates the hidden state of the medical notes and code description; (2) Code Description Aware Attention which computes the label-specific document representation; (3) Associated and Hierarchical Code Description Distillation which distills the noisy medical notes with the aid of the associated codes and their sibling codes' description; (4) Code Description Aware Output which predicts the medical codes.

\subsection{Encoder Layer}
Initially, we transform each word into a compact, low-dimensional word embedding. Specifically, each token \( t_i \) in a document \(d\) is represented by a corresponding pre-trained word embedding \( x_i \), each sharing a uniform embedding size.

In terms of the encoder, theoretically, it can be implemented using any type of neural encoder, such as CNN based encoders \cite{caml,multirescnn}, RNN based encoders \cite{LAAT,MSMN} or Transformer based encoders \cite{Fusion}. For a given medical note embeddings \( X_d = \{x_1,x_2,\ldots,x_{N_d}\} \), the encoder generates a hidden state for each word. The encoder processes these word embeddings to produce text hidden representations \( H_d =\{h_1,h_2,\) 
\( \ldots,h_{N_d}\} \in \mathbb{R}^{N_d \times h} \), where \( h \) denotes the dimensionality of the hidden state.
\begin{equation} 
H_d = \text{Enc}(x_1,x_2,\ldots,x_{N_d}) 
\end{equation}

\subsection{Code Description Aware Attention}
In medical notes, not every word contributes equally to medical diagnosis decision-making. Thus, attention weights are employed to refine medical note representations in alignment with code description representations.

Previous studies \cite{multirescnn,LAAT} have used \( H_d \) as input to compute \( N_L \) label-specific representations. This process involves two steps. Initially, a label-wise attention weight matrix \( \alpha \) is calculated as follows:
\begin{equation} 
\text{Attention}(Q,K,V) = \text{softmax}(QK^T)V 
\end{equation}
\begin{equation} 
\alpha=\text{Attention}(U,H_d,H_d)
\end{equation}
This phase involves learning code-dependent parameters \( U \in \mathbb{R}^{N_L \times h} \) for the attention weights. However, the attention for rare codes may not be effectively learned due to limited training data. To address this, we compute the value of each code description's hidden representations using the same encoder, followed by a max-pooling operation to generate the code's representations \( H^c_i \). Subsequently, we concatenate these code's representations to form \( H_C \). This matrix \( H_C \) undergoes a linear projection to substitute for \( U \). The only parameters required for learning are the linear projection parameters \( W_Q \in \mathbb{R}^{h \times h} \).
\begin{equation} 
H^c_i =\text{MaxPool}(\text{Enc}(x^c_1,x^c_2,\ldots,x^c_{N_{c_i}}))
\end{equation}
\begin{equation} 
H_C = \{H^c_1, H^c_2, \ldots, H^c_{N_l} \}
\end{equation}
\begin{equation} 
{\alpha}_C=\text{Attention}({H_C}{W_Q},H_d,H_d)
\end{equation}
Then, the matrix \( \alpha_C \) is applied to calculate a weighted sum of \( H_d \), creating the label-specific document representation:
\begin{equation} 
V_d = H_d \cdot \alpha_C
\end{equation}

\subsection{Associated and Hierarchical Code Description Distillation}
To deal with the long and noisy text in ICD coding task, we use the associated codes' description to distill the medical note, effectively sifting through and removing irrelevant words from the medical note. Furthermore, to ensure the capture of medical terms that are specifically relevant to a given label and distinct from similar sibling codes, we incorporate the descriptions of these sibling codes in our distillation. This distillation is crucial in removing words that are more closely related to the sibling codes rather than the associated code.

\subsubsection{Associated Code Description Distillation}
Given a medical note \( \mathbf{d} \) and its associated code set \( L_A \), we consider the set of these associated codes' description \( c_A = \{c_1, c_2, \ldots, c_{N_{L_A}}\} \) as critical for classification. Here, \( N_{L_A} \) represents the size of the associated code set \( L_A \). We aim to 
get the information relevant to the associated code by distilling the medical note. Specially, we align the label-specific representations of the medical note closer to those of the associated codes' descriptions, treating the latter as training data for classification.

The label-specific document representation \( V_{c_A} \) of the associated codes' descriptions is obtained using the same encoder and attention layer. We measure the similarity between \( V_{c_A} \) and \( V_d \) using cosine distance, striving to minimize this distance.
\begin{equation} 
\mathrm{cos}(V_d, V_{c_A}) = \frac{1}{N_L}\sum_{i=1}^{N_L} \frac{V^i_d \cdot V^i_{c_A}}{\|V^i_d\|\ \|V^i_{c_A}\|}
\end{equation}
\begin{equation} 
L_{sim} = 1 - \mathrm{cos}(V_d, V_{c_A})
\end{equation}
where, \( V^i_d \) denotes the document representations for the \( i_{\text{th}} \) label.

The loss for associated code descriptions distillation comprises two components: the similarity between \( V_{c_A} \) and \( V_d \), and the binary cross-entropy loss using associated codes' descriptions as training data.

\subsubsection{Hierarchical Code Description Distillation}
ICD codes are structured in a tree-like hierarchy, reflecting parent-child and sibling relationships. Upper-level nodes signify broader disease categories, while lower-level nodes detail specific diseases. This hierarchy helps identify mutually exclusive codes. For instance, if codes X and Y are siblings under Z, it's improbable for both to be assigned to a patient concurrently. However, the sibling codes are more clinically relevant. 

To effectively harness this hierarchical structure and ensure the accurate capture of medical terms that are relevant to a specific label and distinguishable from those associated with similar sibling codes, we utilize a distillation by the hierarchical code description. More specifically, we aim to differentiate the label-specific representation within a medical note from the descriptions pertinent to its sibling codes. 

The loss for hierarchical code description distillation is the dissimilarity between the label-specific document representation of medical note \( V_d \) and sibling codes' descriptions \( V_{c_S} \), striving to maximize the dissimilarity. The dissimilarity between \( V_d \) and \( V_{c_S} \) is quantified similarly to the aforementioned similarity measure:
\begin{equation} 
\mathrm{cos}(V_d, V_{c_S}) = \frac{1}{N_L}\sum_{i=1}^{N_L} \frac{V^i_d \cdot V^i_{c_S}}{\|V^i_d\|\ \|V^i_{c_S}\|}
\end{equation}
\begin{equation} 
L_{dis} = \mathrm{cos}(V_d, V_{c_S})
\end{equation}
where, \( c_S \) denotes the description of sibling codes, and \( V_{c_S} \) represents the label-specific representation of the sibling codes' description \( c_S \).

\subsection{Code Description Aware Output}
For each label-specific representation \( V_d \), we determine whether the medical note \( \mathbf{d} \) contains the code \( l_i \) by evaluating the similarity between the code's and medical note's representations.

In previous works \cite{multirescnn,Fusion}, the label-specific representation \( V_d \) is fed into a linear layer, followed by a sum-pooling operation to generate the score vector \( \hat{y}^d \) for all ICD codes. Subsequently, the probability vector \( \Bar{y}^d \) is derived using the sigmoid function.

\begin{equation} 
\hat{y}^d = \mathrm{SumPool}(V_dW)
\end{equation}
\begin{equation} 
\Bar{y}^d = \mathrm{sigmoid}(\hat{y}^d)
\end{equation}
where, the code-dependent parameter matrix \( W \in \mathbb{R}^{h \times N_L} \) must be learned for classification. However, due to imbalanced data distribution, learning the weight matrix \( W \) can be challenging. To address this, we have incorporate the similarity between the code’s and medical note’s representations. Specially, we calculate \( W^C \), derived from all code descriptions’ hidden representations, and then passed through a linear layer. Then we use \( V_dW^C + V_dW \) in place of the original \( V_dW \) to focus more intensely on those rare codes. 
\begin{equation} 
W^C = (H^CW^l)^T, \quad W^l \in \mathbb{R}^{h \times h}
\end{equation}
\begin{equation} 
\hat{y}^d = \mathrm{SumPool}(V_dW^C + V_dW)
\end{equation}

\subsection{Training}
The training objective is to minimize the binary cross-entropy loss between the prediction \( \Bar{y}^d \) and the target \( y \), using the medical note \( d \) as training data. Additionally, we incorporate associated and hierarchical code descriptions to distill the medical note, capturing relevant medical words and the code hierarchy.
\begin{equation} 
\mathrm{Loss}(d, y) = -\sum_{i=1}^{N_L} (y_i\mathrm{log}\Bar{y}^d_i + (1-y_i)(1-\mathrm{log}\Bar{y}^d_i))
\end{equation}
\begin{equation} 
L = \mathrm{Loss}(d, y) + \lambda_{sim} \cdot L_{sim} + \mathrm{Loss}(c_A, y) + \lambda_{dis} \cdot L_{dis}
\end{equation}
\section{Experiments}
\subsection{Datasets}
In our study, we utilize the widely accessible MIMIC-III dataset \cite{Mimic3_dataset}. The dataset encompasses 8,921 unique ICD-9 codes, subdivided into 6,918 diagnosis codes and 2,003 procedure codes. In alignment with earlier studies and to ensure uniform distribution of a patient's notes across the training, validation, and test sets, we adopt the data split methodology proposed by \cite{caml}. The MIMIC-III dataset offers two primary configurations: the comprehensive MIMIC-III Full and the more focused MIMIC-III 50. The former includes all 8,921 codes and contains 47,719, 1,631, and 3,372 discharge summaries for training, development, and testing, respectively. The latter, MIMIC-III 50, encompasses the top 50 most frequent codes and includes 8,067, 1,574, and 1,730 discharge summaries for the respective phases. Our evaluation approach is informed by the methodology of \cite{MSMN}, employing Micro and Macro AUC (area under the ROC curve), Micro and Macro F1 scores, and Precision@K as our primary metrics. For the MIMIC-III 50 dataset, we specifically report Precision@5 (P@5) and, for the comprehensive MIMIC-III Full, Precision@8 (P@8).

\subsection{Baselines}
We compare our method with the following baselines:

\textbf{CAML} \cite{caml}: A pioneering model in automated ICD coding, utilizing a label attention layer for generating label-specific representations.

\textbf{MultiResCNN} \cite{multirescnn}: This model leverages a multi-filter CNN to capture diverse patterns in medical notes and a residual block to expand the model’s receptive field.

\textbf{LAAT} \cite{LAAT}: This model proposes a new label attention model to learn attention scores over BiLSTM encoding hidden states for each medical code.

\textbf{Fusion} \cite{Fusion}: This model combines multi-CNN, Transformer encoder, and label attention to enhance performance and accuracy in ICD coding.

\textbf{MSMN} \cite{MSMN}:This model uses code synonyms collected from medical knowledge graph with adapted multi-head attention and LSTM encoder.

\textbf{Rare-ICD} \cite{2023Rare-ICD}:This model uses relations between different codes via a relationenhanced code encoder to improve the rare code performance.

\subsection{Results}

\begin{table*}
\centering
\caption{Results on the MIMIC-III Full and MIMIC-III 50 datasets.}
\begin{tblr}{
  width = \linewidth,
  colspec = {Q[200]Q[70]Q[70]Q[70]Q[70]Q[70]Q[70]Q[70]Q[70]Q[70]Q[70]},
  row{2} = {c},
  row{3} = {c},
  cell{1}{1} = {r=3}{0.1\linewidth},
  cell{1}{2} = {c=5}{0.3\linewidth,c},
  cell{1}{7} = {c=5}{0.3\linewidth,c},
  cell{2}{2} = {c=2}{0.12\linewidth},
  cell{2}{4} = {c=2}{0.12\linewidth},
  cell{2}{6} = {r=2}{},
  cell{2}{7} = {c=2}{0.12\linewidth},
  cell{2}{9} = {c=2}{0.12\linewidth},
  cell{2}{11} = {r=2}{},
  cell{4}{2} = {c},
  cell{4}{3} = {c},
  cell{4}{4} = {c},
  cell{4}{5} = {c},
  cell{4}{6} = {c},
  cell{4}{7} = {c},
  cell{4}{8} = {c},
  cell{4}{9} = {c},
  cell{4}{10} = {c},
  cell{4}{11} = {c},
  cell{5}{2} = {c},
  cell{5}{3} = {c},
  cell{5}{4} = {c},
  cell{5}{5} = {c},
  cell{5}{6} = {c},
  cell{5}{7} = {c},
  cell{5}{8} = {c},
  cell{5}{9} = {c},
  cell{5}{10} = {c},
  cell{5}{11} = {c},
  cell{6}{2} = {c},
  cell{6}{3} = {c},
  cell{6}{4} = {c},
  cell{6}{5} = {c},
  cell{6}{6} = {c},
  cell{6}{7} = {c},
  cell{6}{8} = {c},
  cell{6}{9} = {c},
  cell{6}{10} = {c},
  cell{6}{11} = {c},
  cell{7}{2} = {c},
  cell{7}{3} = {c},
  cell{7}{4} = {c},
  cell{7}{5} = {c},
  cell{7}{6} = {c},
  cell{7}{7} = {c},
  cell{7}{8} = {c},
  cell{7}{9} = {c},
  cell{7}{10} = {c},
  cell{7}{11} = {c},
  cell{8}{2} = {c},
  cell{8}{3} = {c},
  cell{8}{4} = {c},
  cell{8}{5} = {c},
  cell{8}{6} = {c},
  cell{8}{7} = {c},
  cell{8}{8} = {c},
  cell{8}{9} = {c},
  cell{8}{10} = {c},
  cell{8}{11} = {c},
  cell{9}{2} = {c},
  cell{9}{3} = {c},
  cell{9}{4} = {c},
  cell{9}{5} = {c},
  cell{9}{6} = {c},
  cell{9}{7} = {c},
  cell{9}{8} = {c},
  cell{9}{9} = {c},
  cell{9}{10} = {c},
  cell{9}{11} = {c},
  cell{10}{2} = {c},
  cell{10}{3} = {c},
  cell{10}{4} = {c},
  cell{10}{5} = {c},
  cell{10}{6} = {c},
  cell{10}{7} = {c},
  cell{10}{8} = {c},
  cell{10}{9} = {c},
  cell{10}{10} = {c},
  cell{10}{11} = {c},
  cell{11}{2} = {c},
  cell{11}{3} = {c},
  cell{11}{4} = {c},
  cell{11}{5} = {c},
  cell{11}{6} = {c},
  cell{11}{7} = {c},
  cell{11}{8} = {c},
  cell{11}{9} = {c},
  cell{11}{10} = {c},
  cell{11}{11} = {c},
  cell{12}{2} = {c},
  cell{12}{3} = {c},
  cell{12}{4} = {c},
  cell{12}{5} = {c},
  cell{12}{6} = {c},
  cell{12}{7} = {c},
  cell{12}{8} = {c},
  cell{12}{9} = {c},
  cell{12}{10} = {c},
  cell{12}{11} = {c},
  cell{13}{2} = {c},
  cell{13}{3} = {c},
  cell{13}{4} = {c},
  cell{13}{5} = {c},
  cell{13}{6} = {c},
  cell{13}{7} = {c},
  cell{13}{8} = {c},
  cell{13}{9} = {c},
  cell{13}{10} = {c},
  cell{13}{11} = {c},
  cell{14}{2} = {c},
  cell{14}{3} = {c},
  cell{14}{4} = {c},
  cell{14}{5} = {c},
  cell{14}{6} = {c},
  cell{14}{7} = {c},
  cell{14}{8} = {c},
  cell{14}{9} = {c},
  cell{14}{10} = {c},
  cell{14}{11} = {c},
  cell{15}{2} = {c},
  cell{15}{3} = {c},
  cell{15}{4} = {c},
  cell{15}{5} = {c},
  cell{15}{6} = {c},
  cell{15}{7} = {c},
  cell{15}{8} = {c},
  cell{15}{9} = {c},
  cell{15}{10} = {c},
  cell{15}{11} = {c},
  vline{2,7} = {1-15}{},
  hline{1,16} = {-}{0.08em},
  hline{3} = {2-5,7-10}{},
  hline{4,6,8,10,12,14,16} = {-}{},
}
Model& \textbf{MIMIC-III Full} & & & & & \textbf{MIMIC-III 50} & & & & \\
& Auc & & F1 & & P@8 & Auc &  & F1 & & P@5 \\
& Macro & Micro & Macro & Micro & & Macro & Micro & Macro & Micro & \\
CAML        
& 88.0~ & 98.3~ & 5.7~ & 50.2~ & 69.8~ 
& 87.3~ & 90.6~ & 49.6~ & 60.3~ & 60.5~ \\
+AHDD       
& \textbf{90.5~} & \textbf{98.5~} & \textbf{5.8~} & \textbf{50.7~} & \textbf{69.8}~
& \textbf{89.1~} & \textbf{91.6~} & \textbf{56.3~} & \textbf{63.1~} & \textbf{62.0}~ \\

MultiResCNN 
& 90.5~ & 98.6~ & 7.6~ & 55.1~ & 73.8~ 
& 89.7~ & 92.5~ & 59.8~ & 66.8~ & 63.3~ \\
+AHDD       
& \textbf{90.6~} & \textbf{98.6~} & \textbf{8.6~} & \textbf{56.2~} & \textbf{73.9}~
& \textbf{90.7~} & \textbf{93.2~} & \textbf{61.3~} & \textbf{67.6~} & \textbf{64.0}~ \\

LAAT       
& 91.9~ & 98.8~ & 9.9~ & 57.5~ & 73.8~ 
& 92.5~ & 94.6~ & 66.1~ & 71.6~ & 67.1~ \\
+AHDD       
& \textbf{94.7~} & \textbf{99.1~} & \textbf{10.2~} & \textbf{57.9~} & \textbf{74.7~} 
& \textbf{92.5~} & \textbf{94.6~} & \textbf{67.8~} & \textbf{71.7~} & \textbf{67.3~}\\ 

Fusion      
& 91.5~  & 98.7~ & 8.3~ & 55.4~ & 73.6~ 
& 90.3~  & 93.1~ & 60.4~ & 67.8~ & 63.8~ \\
+AHDD       
& \textbf{92.3~} & \textbf{98.8}~ & \textbf{9.2~} & \textbf{56.3~} & \textbf{74.1~}
& \textbf{91.2~} & \textbf{93.2}~ & \textbf{62.6~} & \textbf{68.1~} & \textbf{64.4~} \\

MSMN        
& 95.0~ & 99.2~ & 10.3~ & 58.2~ & 74.9~ 
& 92.7~ & 94.6~ & 67.4~ & 71.7~ & 67.4~ \\
+AHDD       
& \textbf{95.0~} & \textbf{99.2~} & \textbf{10.4~} & \textbf{58.8~} & \textbf{75.3~} 
& \textbf{92.8~} & \textbf{94.7~} & \textbf{68.5~} & \textbf{72.8~} & \textbf{67.8~}\\

Rare-ICD     
& 94.7~  & 99.1~ & 10.5~ & 58.1~ & 74.5~  
& 91.9~  & 94.2~ & 64.2~ & 70.8~ & 66.5~ \\
+AHDD       
& \textbf{95.2~} & \textbf{99.3}~ & \textbf{10.9~} & \textbf{58.9~} & \textbf{75.3~}
& \textbf{92.1~} & \textbf{94.3}~ & \textbf{65.8~} & \textbf{71.3~} & \textbf{66.8~} \\
\end{tblr}
\label{Table_50}
\end{table*}

\begin{figure*}[h]
    \centering
    \begin{minipage}[b]{\linewidth}
        \centering
        \includegraphics[width=.3\linewidth]{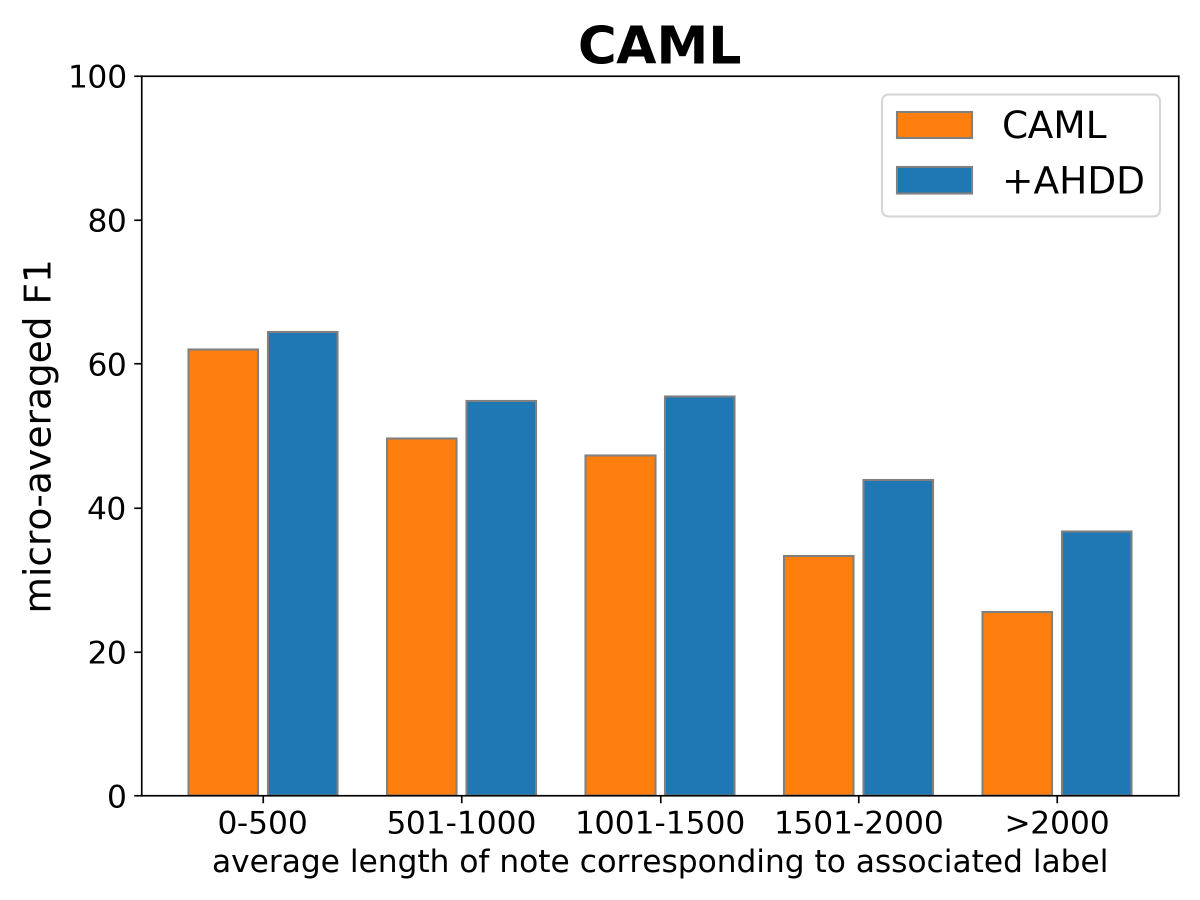}
        \includegraphics[width=.3\linewidth]{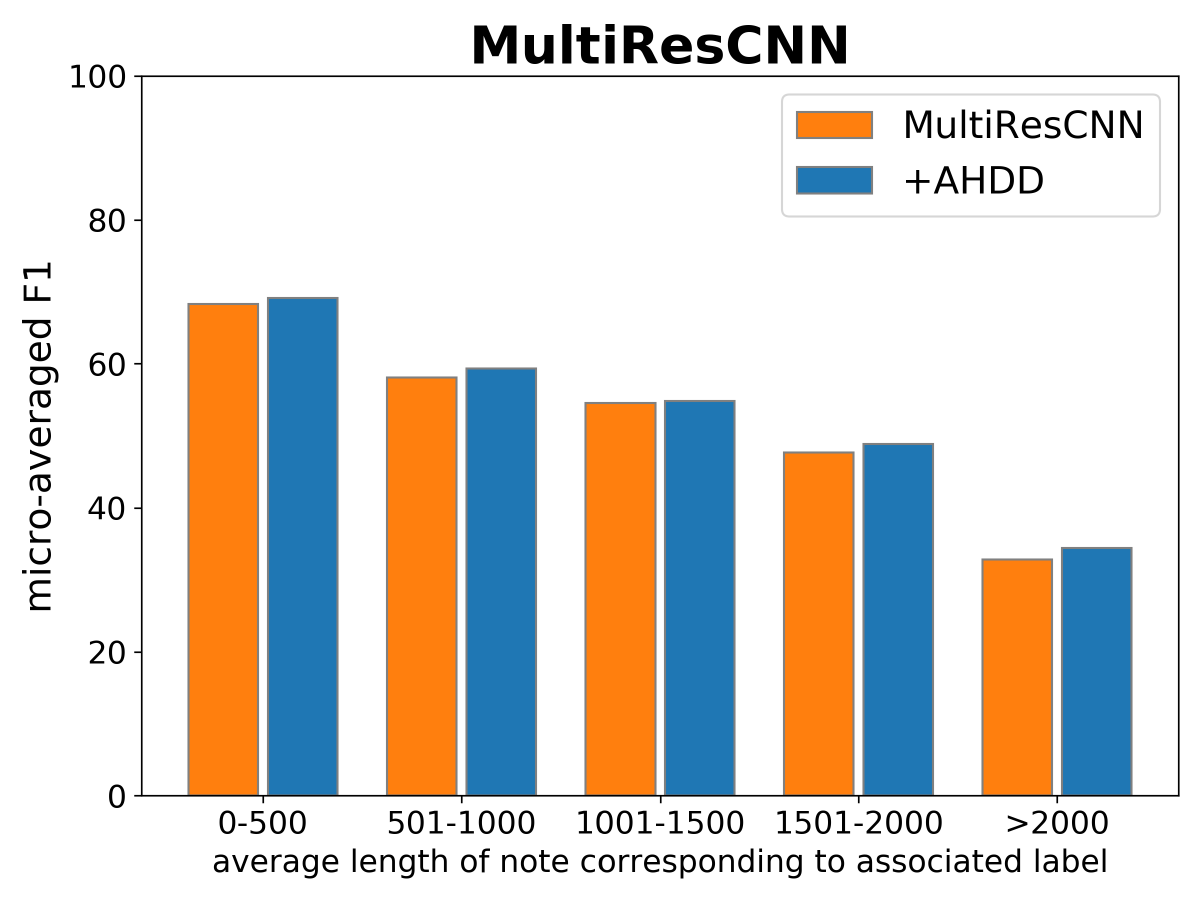}
        \includegraphics[width=.3\linewidth]{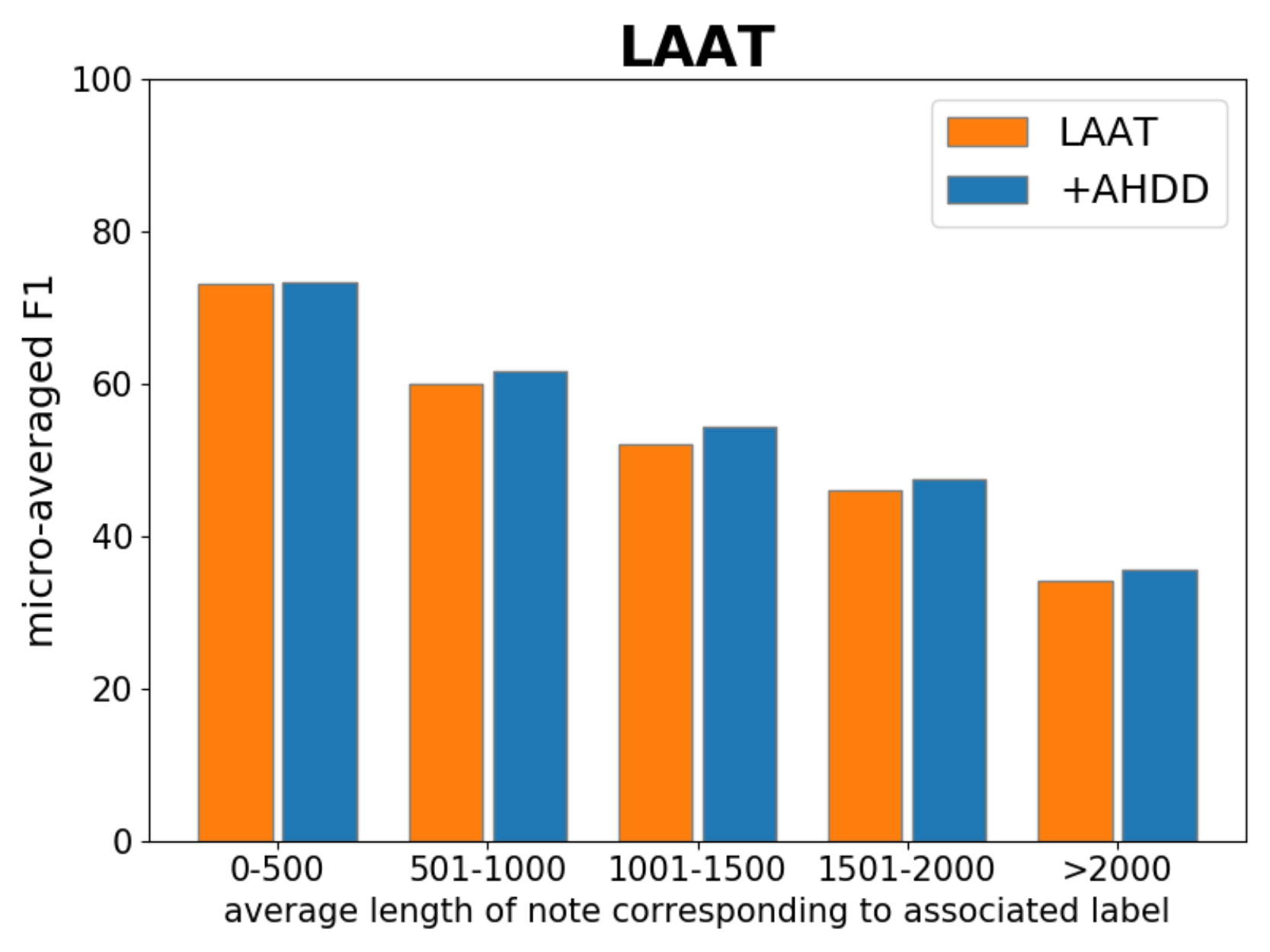}
    \end{minipage}
    \centering
    \begin{minipage}[b]{\linewidth}
        \centering
        \includegraphics[width=.3\linewidth]{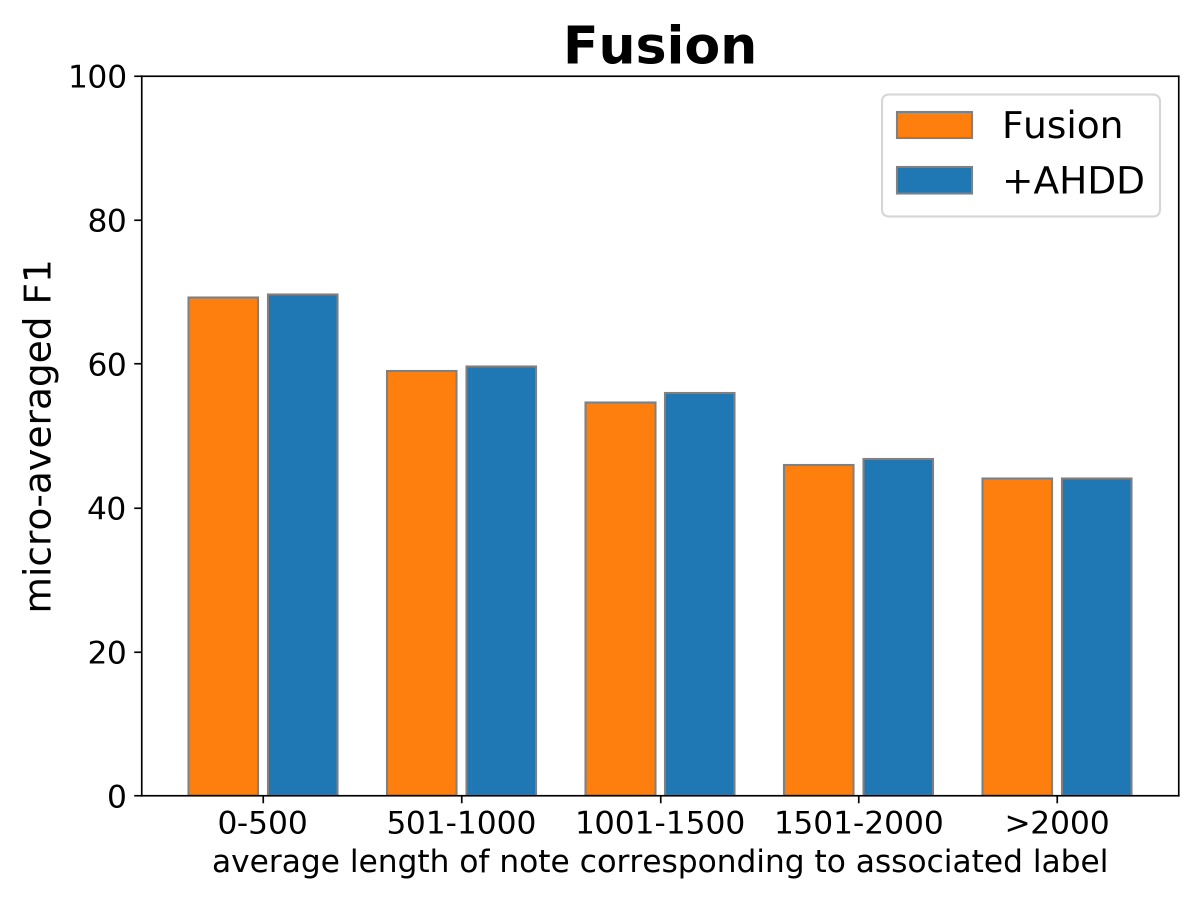}
        \includegraphics[width=.3\linewidth]{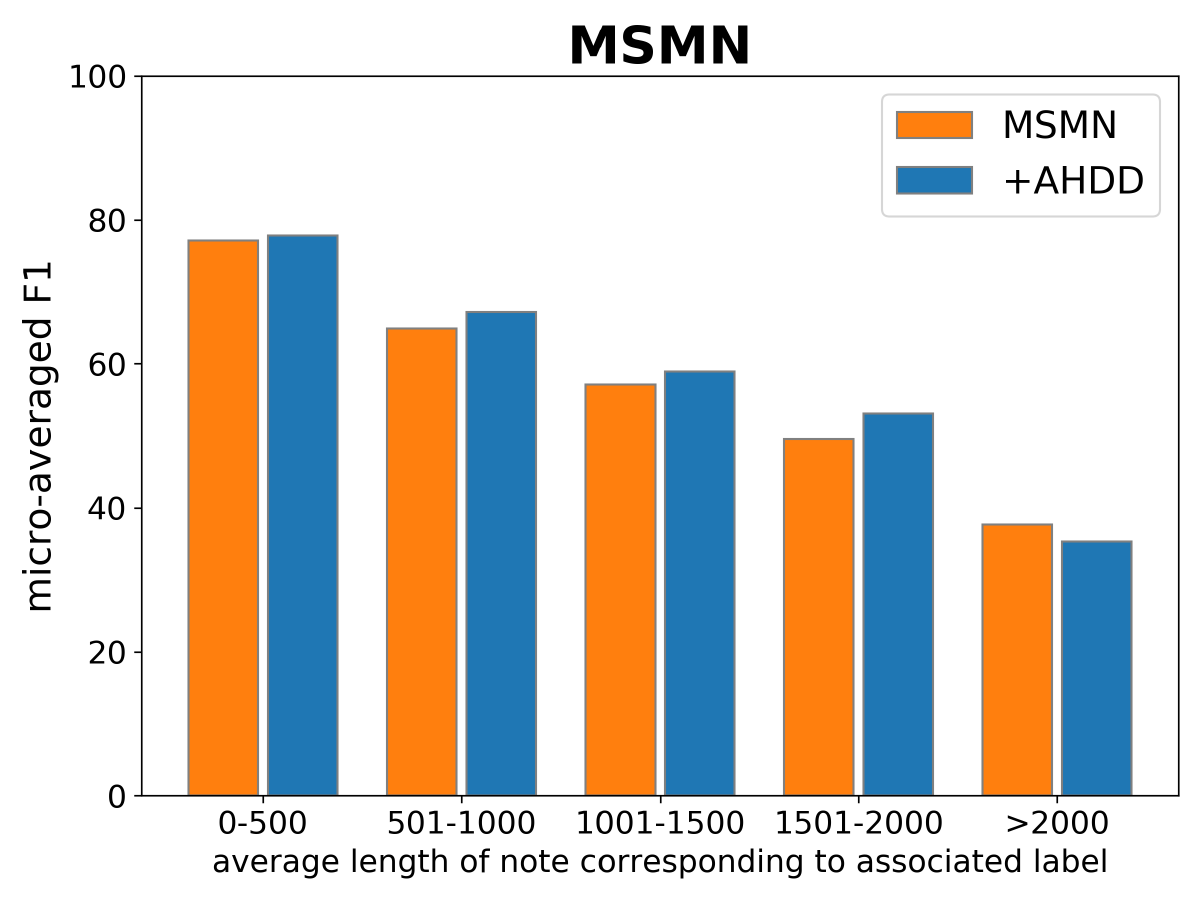}
        \includegraphics[width=.3\linewidth]{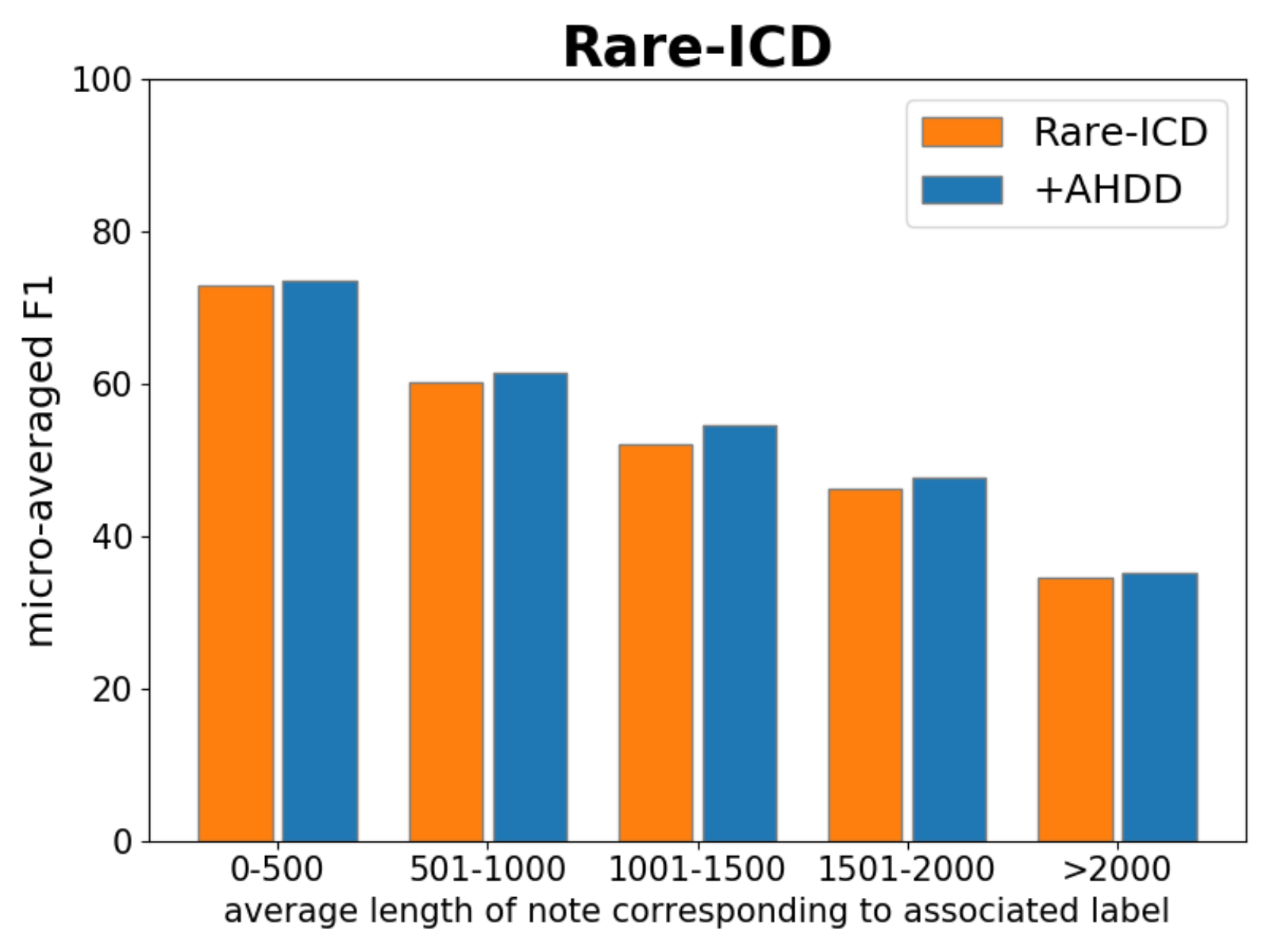}
    \end{minipage}

    \caption{Micro-averaged F1 for the average note length groups associated with label for CAML, MultiResCNN, LAAT, Fusion, MSMN and Rare-ICD models. The x-axis represents the groups based on average note length, while the y-axis shows the micro-averaged F1 for each of these groups. }
	\label{Figure 3}
\end{figure*}

Table \ref{Table_50} presents a comparison of performance between the baseline models and the proposed AHDD method under both the MIMIC-III Full and MIMIC-III 50 settings.

Overall, it can be observed that the implementation of the AHDD method enhances the performance of all baseline models. This demonstrates the effectiveness and necessity of distilling medical notes by associated and hierarchical code description. The significant improvement in F1 and Precision@K scores further substantiate the efficacy of the proposed method in capturing relevant medical words for clinicians. Performance metrics, particularly those focusing on the micro scale, provide insight into the effectiveness of addressing the issue of noisy text. 

To further validate this deduction, we have categorized medical notes into five distinct groups based on the average note length associated with label: [0,500], [501,1000], [1001,1500], [1501,2000], and [2001, +\( \infty \)]. For each of these average note length groups, we computed the micro-averaged F1 score across all baseline methods and compared them to their counterparts within our AHDD method. The summarized results are presented in \figref{Figure 3}.

A notable observation from this analysis is the significant improvement in F1 scores that AHDD brings to the [0,500], [501,1000], [1001,1500], and [1501,2000] groups. Furthermore, these relative improvements become increasingly pronounced with the lengthening of the average medical note. This trend underscores the effectiveness of our AHDD method in identifying key words within noisy and complex medical notes, validating its capability to handle such challenging data effectively.



\subsection{Ablation Study}
\begin{table}
\centering
\caption{Ablation results on the MIMIC-III 50 dataset. }
\begin{tblr}{
  width = \linewidth,
  colspec = {Q[369]Q[117]Q[106]Q[117]Q[106]Q[98]},
  cells = {c},
  cell{1}{1} = {r=2}{},
  cell{1}{2} = {c=2}{0.223\linewidth},
  cell{1}{4} = {c=2}{0.223\linewidth},
  cell{1}{6} = {r=2}{},
  hline{1-8} = {-}{},
  hline{2} = {2-6}{},
}
Model                 & Auc   &       & F1    &       & P@5   \\
                      & Macro & Micro & Macro & Micro &       \\
MSMN+AHDD             & 92.8~ & 94.7~ & 68.5~ & 72.8~ & 67.8~ \\
w/o ADD               & 92.8~ & 94.7~ & 68.0~ & 72.2~ & 67.5~ \\
w/o HDD~              & 92.8~ & 94.6~ & 67.9~ & 71.9~ & 67.8~ \\
w/o D-att & 92.8~ & 94.7~ & 68.5~ & 72.3~ &67.7~ \\
w/o D-output & 92.8~ & 94.7~ & 68.0~ & 72.0~ &67.6~
\end{tblr}
\label{Table ablation} 
\end{table}
To comprehensively assess the impact of each individual component, we carried out a series of ablation experiments on MSMN+AHDD. The results of these experiments are detailed in Table \ref{Table ablation}. From these results, it can be observed that:

When discarding the associated code description distillation (w/o ADD in Table \ref{Table ablation}), the performance drops dramatically in various metrics, especially on P@5 metric. It shows the effectiveness of associated code description distillation in capturing key words in noisy and complex medical notes.

When discarding the hierarchical code description distillation (w/o HDD in Table \ref{Table ablation}), the performance drops obviously in various metrics, especially on F1 metric. But the metric of P@5 keeps the same. It shows the effectiveness of hierarchical code description distillation in avoidance of inappropriate code assignments.

Without Code Description Aware Attention (w/o D-att in Table \ref{Table ablation}), attention layers degrades to attention layers in previous works. The performance drops obviously in various metrics, especially on F1 and P@5 metric. It shows a significant reduction in the ability to capture important words.

Without Code Description Aware Output (w/o D-output in Table \ref{Table ablation}),  output layers degrades to output layers in previous works. The performance drops obviously in various metrics, especially on F1 and P@5 metric. It shows a significant reduction in the ability to predict rare codes.
\subsection{Case Study}
\begin{figure}[h]
	\centering
	\includegraphics[scale=0.55]{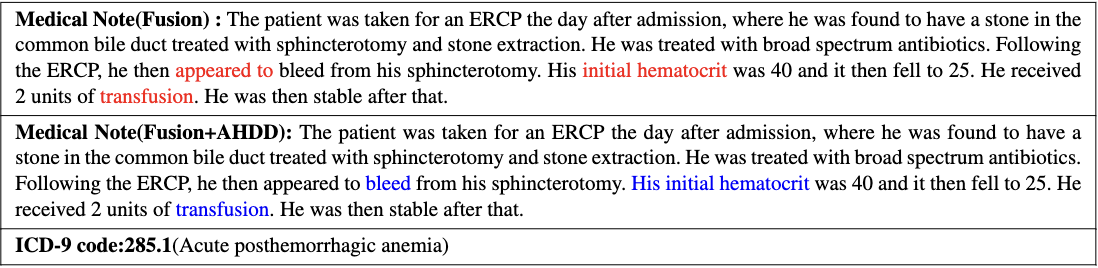} 
	\caption{The attention distribution visualization over a medical note with a medizcal code for Fusion and its counterparts under AHDD method. We highlight the highly weighted words}
	\label{Figure visual}
\end{figure}
To further explore the proposed method's ability to capture relevant information, we present a visualization of the attention distribution over a medical note. This visualization compares Fusion and its counterparts within our AHDD method, as illustrated in \figref{Figure visual}.

It can be observed that Fusion only captures scattered label-related words like "\textit{appeared to}", "\textit{initial hematocrit}" and "\textit{transfusion}" for inferring "\textit{anemia}", while it fails to find valid relevant evidence for inferring "\textit{Acute posthemorrhagic anemia}". In contrast, the AHDD method, which utilizes associated and hierarchical code description distillation, successfully identifies the crucial word "\textit{bleed}" to infer ICD code 285.1. Therefore, through the distillation , the model can more effectively capture significant words and avoid improper code assignments.
\section{Conclusion}
We propose a novel method based on associated and hierarchical code description distillation to predict ICD codes. The code description is also applied to aware the attention layer and output layer. To the best of our knowledge, our work is the first attempt to use associated and hierarchical code descriptions for distilling medical notes. The method is encoder-agnostic, and it neither needs extra text processing nor brings in too many parameters. The experimental results demonstrate that our proposed method surpasses several leading state-of-the-art models on a benchmark dataset.

\bigskip
\noindent\textbf{Acknowledgements.} This work was supported by the National Key Research and Development Program of China under Grant 2023YFB3002201.


%
%
%
\bibliography{reference}

\begin{thebibliography}{10}
\providecommand{\url}[1]{\texttt{#1}}
\providecommand{\urlprefix}{URL }
\providecommand{\doi}[1]{https://doi.org/#1}

\bibitem{transicd}
Biswas, B., Pham, T.H., Zhang, P.: Transicd: Transformer based code-wise attention model for explainable icd coding. In: Artificial Intelligence in Medicine. pp. 469--478. Springer International Publishing, Cham (2021). \doi{10.1007/978-3-030-77211-6_56}

\bibitem{hypercore}
Cao, P., Chen, Y., Liu, K., Zhao, J., Liu, S., Chong, W.: Hypercore: Hyperbolic and co-graph representation for automatic icd coding. In: ACL (Jan 2020)

\bibitem{2023Rare-ICD}
Chen, J., Li, X., Xi, J., Yu, L., Xiong, H.: Rare codes count: Mining inter-code relations for long-tail clinical text classification. In: Clinical Natural Language Processing Workshop (2023). \doi{10.18653/v1/2023.clinicalnlp-1.43}

\bibitem{monitor}
Choi, E., Bahadori, M.T., Schuetz, A., Stewart, W.F., Sun, J.: Doctor ai: Predicting clinical events via recurrent neural networks. In: Proceedings of Machine Learning Research. vol.~56, pp. 301--318. PMLR (18--19 Aug 2016)

\bibitem{2024AccurateAW}
Gomes, G., Coutinho, I., Martins, B.: Accurate and well-calibrated icd code assignment through attention over diverse label embeddings. In: EACL (2024)

\bibitem{Mimic3_dataset}
Johnson, A.E., Pollard, T.J., Shen, L., Lehman, L.w.H., Feng, M., Ghassemi, M., Moody, B., Szolovits, P., Anthony~Celi, L., Mark, R.G.: Mimic-iii, a freely accessible critical care database. Scientific Data  (May 2016). \doi{10.1038/sdata.2016.35}

\bibitem{multirescnn}
Li, F., Yu, H.: Icd coding from clinical text using multi-filter residual convolutional neural network. Proceedings of the AAAI Conference on Artificial Intelligence p. 8180–8187 (Jun 2020). \doi{10.1609/aaai.v34i05.6331}

\bibitem{2024CoRelationBA}
Luo, J., Wang, X., Wang, J., Chang, A., Wang, Y., Ma, F.: Corelation: Boosting automatic icd coding through contextualized code relation learning. ArXiv  (2024)

\bibitem{Fusion}
Luo, J., Xiao, C., Glass, L., Sun, J., Ma, F.: Fusion: Towards automated icd coding via feature compression. In: Findings of the Association for Computational Linguistics: ACL-IJCNLP 2021 (Jan 2021). \doi{10.18653/v1/2021.findings-acl.184}

\bibitem{rule}
Medori, J., Fairon, C.: Machine learning and features selection for semi-automatic {ICD}-9-{CM} encoding. In: Proceedings of the {NAACL} {HLT} 2010 Second Louhi Workshop on Text and Data Mining of Health Documents. pp. 84--89 (Jun 2010)

\bibitem{caml}
Mullenbach, J., Wiegreffe, S., Duke, J., Sun, J., Eisenstein, J.: Explainable prediction of medical codes from clinical text. In: NAACL (Jan 2018)

\bibitem{manual}
O’Malley, K.J., Cook, K.F., Price, M.D., Wildes, K.R., Hurdle, J.F., Ashton, C.M.: Measuring diagnoses: Icd code accuracy. Health Services Research  \textbf{40}(5p2),  1620–1639 (Oct 2005)

\bibitem{svm}
Perotte, A., Pivovarov, R., Natarajan, K., Weiskopf, N., Wood, F., Elhadad, N.: Diagnosis code assignment: models and evaluation metrics. Journal of the American Medical Informatics Association p. 231–237 (Mar 2014). \doi{10.1136/amiajnl-2013-002159}

\bibitem{decision}
Scheurwegs, E., Cule, B., Luyckx, K., Luyten, L., Daelemans, W.: Selecting relevant features from the electronic health record for clinical code prediction. Journal of Biomedical Informatics  \textbf{74},  92–103 (Oct 2017). \doi{10.1016/j.jbi.2017.09.004}

\bibitem{epidemiological}
Tsui, F.C.: Value of icd-9-coded chief complaints for detection of epidemics. Journal of the American Medical Informatics Association  \textbf{9}(90061),  41S – 47 (Nov 2002)

\bibitem{LAAT}
Vu, T., Nguyen, D.Q., Nguyen, A.: A label attention model for icd coding from clinical text. In: IJCAI (Jul 2020). \doi{10.24963/ijcai.2020/461}

\bibitem{2023KnowledgeMKFNMK}
Wang, S., Lin, H., Zhang, Y., Li, X., Qu, W.: Mkfn: Multimodal knowledge fusion network for automatic icd coding. 2023 IEEE International Conference on Bioinformatics and Biomedicine (BIBM) pp. 2294--2297 (2023)

\bibitem{knowledge}
Wang, T., Zhang, L., Ye, C., Liu, J., Zhou, D.: A novel framework based on medical concept driven attention for explainable medical code prediction via external knowledge. In: Findings of the Association for Computational Linguistics: ACL 2022. pp. 1407--1416 (May 2022)

\bibitem{2024KnowledgebasedDP}
Xie, J., Li, X., Yuan, Y., Guan, Y., Jiang, J., Guo, X., Peng, X.: Knowledge-based dynamic prompt learning for multi-label disease diagnosis. Knowledge-Based Systems  \textbf{286},  111395 (2024)

\bibitem{hierachy}
Xie, P., Xing, E.: A neural architecture for automated icd coding. In: ACL (Jan 2018)

\bibitem{MSMN}
Yuan, Z., Tan, C., Huang, S.: Code synonyms do matter: Multiple synonyms matching network for automatic {ICD} coding. In: ACL. pp. 808--814 (May 2022)

\bibitem{noisy_distill}
Zhou, T., Cao, P., Chen, Y., Liu, K., Zhao, J., Niu, K., Chong, W., Liu, S.: Automatic icd coding via interactive shared representation networks with self-distillation mechanism. In: ACL (Jan 2021). \doi{10.18653/v1/2021.acl-long.463}

\end{thebibliography}
\bibliographystyle{splncs04}

\end{document}